\documentclass[11pt]{article}
\usepackage{acl2012} 

\usepackage{epsfig}
\usepackage{epstopdf}
\usepackage{times}
\usepackage{graphicx}
\usepackage[round]{natbib}
\usepackage[hyphens]{url}
\usepackage{paralist}
\usepackage{slashbox}

\usepackage{booktabs}
\usepackage[table,xcdraw]{xcolor}
\usepackage{dblfloatfix}
\usepackage{amsmath}

\newcommand{\textnl}{\textsl}

\usepackage{color}

\newcommand{\mi}{m_{i}}
\newcommand{\splitcellleft}[2][l]{%
\begin{tabular}[#1]{@{}l@{}}#2\end{tabular}}
\newcommand{\splitcellcen}[2][c]{%
\begin{tabular}[#1]{@{}c@{}}#2\end{tabular}}

\DeclareMathOperator*{\argmin}{arg\,min}

\makeatletter
\newcommand{\@BIBLABEL}{\@emptybiblabel}
\newcommand{\@emptybiblabel}[1]{}
\makeatother
\usepackage[hidelinks]{hyperref}

\title{Unsupervised Identification of Translationese}
\author{
\begin{tabular}[t]{c@{\extracolsep{4em}}c}
Ella Rabinovich & Shuly Wintner \\
\textnormal{Department of Computer Science} & \textnormal{Department of Computer Science} \\
\textnormal{University of Haifa} & \textnormal{University of Haifa} \\
\textnormal{\url{ellarabi@csweb.haifa.ac.il}} & \textnormal{\url{shuly@cs.haifa.ac.il}} \\
\end{tabular}
}

\date{}

\begin{document}
\maketitle
\thispagestyle{empty}


\begin{abstract}
Translated texts are distinctively different from original ones, to the extent that supervised text classification methods can distinguish between them with high accuracy. These differences were proven useful for statistical machine translation. However, it has been suggested that the accuracy of translation detection deteriorates when the classifier is evaluated outside the domain it was trained on.
We show that this is indeed the case, in a variety of evaluation scenarios. We then show that \emph{unsupervised} classification is highly accurate on this task. We suggest a method for determining the correct labels of the clustering outcomes, and then use the labels for voting, improving the accuracy even further. Moreover, we suggest a simple method for clustering in the challenging case of mixed-domain datasets, in spite of the dominance of domain-related features over translation-related ones. The result is an effective, fully-unsupervised method for distinguishing between original and translated texts that can be applied to new domains with reasonable accuracy.
\end{abstract} 

\section{Introduction}
\label{sec:introduction}
Human-translated texts (in any language) have distinct features that distinguish them from original, non-translated texts.
These differences stem either from the effect of the translation process on the translated outcomes, or from ``fingerprints'' of the source language on the target language product. The term \emph{translationese} was coined to indicate the unique properties of translations.

Awareness to translationese can improve statistical machine translation (SMT). First, for training translation models, parallel texts that were translated in the direction of the SMT task are preferable to texts translated in the opposite direction; second, for training language models, monolingual corpora of translated texts are better than original texts.

It is possible to automatically distinguish between original (O) and translated (T) texts, with very high accuracy, by employing text classification methods. Existing approaches, however, only employ \emph{supervised} machine-learning; they therefore  suffer from two main drawbacks:
\begin{inparaenum}[(i)]
\item they inherently depend on data annotated with the translation direction, and
\item they may not be generalized to unseen (related or unrelated) domains.%
\footnote{We use ``domain'' rather freely henceforth to indicate not only the topic of a corpus but also its modality (written vs.\ spoken), register, genre, date, etc.}
\end{inparaenum}
These shortcomings undermine the usability of supervised methods for translationese identification in a typical real-life scenario, where no labelled in-domain data are available.

In this work we explore \emph{unsupervised} techniques for reliable discrimination of original and translated texts. More precisely, we apply \emph{dimension reduction} and \emph{centroid-based clustering} methods (enhanced by internal clustering evaluation), for telling O from T in an unsupervised scenario. Furthermore, we introduce a robust methodology for labelling the obtained clusters, i.e., annotating them as ``original'' or ``translated'', by inspecting similarities between the clustering outcomes and O and T \emph{prototypical} examples. Rigorous experiments with four diverse corpora demonstrate that clustering of in-domain texts using lexical, content-independent features systematically yields very high accuracy, only 10 percent points lower than the performance of supervised classification on the same data (in most cases). Accuracy can be improved even further by \emph{clustering consensus} techniques.

We further scrutinize the tension between domain-related and translationese-based text properties. Using a series of experiments in a \emph{mixed-domain} setup, we show that clustering (in particular, relying on content-independent features) perfectly groups the data into domains, rather than into the (desirable) cross-domain O and T; that is, domain-related properties clearly dominate and overshadow the translationese-based characteristics of the underlying texts. We address the challenge of discriminating O from T in a mixed-domain setup by proposing two simple methodologies (\emph{flat} and \emph{two-phase}) and empirically demonstrate their soundness.

The clustering experiments throughout the paper were conducted in a setup similar to that of supervised classification, determining the status (O vs.\ T) of logical units (chunks) of 2,000 tokens. We also show that clustering accuracy remains stable even when the number of available chunks decreases dramatically and remains satisfactory when the chunk size is reduced.

The main contribution of this work is therefore two-fold:
\begin{inparaenum}[(i)]
\item we establish a robust approach for reliable unsupervised identification of translated texts, thereby eliminating the need for in-domain labeled data;
\item we provide an extensive empirical foundation for the dominance of domain-based properties over translationese-related characteristics of a text, and propose a methodology for identification of translationese in a mixed-domain scenario.
\end{inparaenum}

The remainder of the paper is structured as following: after reviewing related work in Section~\ref{sec:related-work}, we detail our datasets in Section~\ref{sec:exp-setup}. In Section~\ref{sec:supervised} we reproduce and extend supervised classification results, and demonstrate the poor cross-domain classification accuracy of supervised methods. Our clustering methodology and experiments are described in Section~\ref{sec:clustering-results}; mixed-domain classification is discussed in Section~\ref{sec:domain-mix}. We conclude with a discussion and suggestions for future research.


\section{Related Work}
\label{sec:related-work}

Much research in Translation Studies indicates that translated texts have unique characteristics. Translated texts (in any language) constitute a sub-language (sometimes referred to as a \emph{genre}, or a \emph{dialect}) of the target language, presumably reflecting both the artifacts of the translation process and traces of the original language from which the texts were translated (the \emph{source} language). \citet{Gellerstam:1986} called this sub-language \emph{translationese}, and suggested that the  differences between O and T do not indicate poor translation but rather a \emph{statistical phenomenon}, caused by a systematic influence of the source language on the target language.

These differences have ramifications for SMT.
\citet{Kurokawa:etal:2009} were the first to note it: they showed that \emph{translation models} trained on English-translated-to-French bitexts were much better than ones trained on French-translated-to-English, when the SMT task is translating English to French. \citet{lembersky-ordan-wintner:2012:EACL2012,lembersky-ordan-wintner:CL2014} corroborated these results, for more language pairs, and suggested a way to adapt translation models to the  properties of translationese. Furthermore, \citet{lembersky-ordan-wintner:2011:EMNLP,lembersky-ordan-wintner:CL2012} showed that \emph{language models} compiled from translated texts are better for SMT than ones compiled from original texts. These results all highlight the practical importance of being able to reliably distinguish between translated and original texts.

Indeed, translated texts are so markedly different from original ones that automatic classification can identify them with very high accuracy \citep{Baroni2006,Ilisei2010,Ilisei:2012,Popescu11}. Recently, \cite{vered:noam:shuly} investigated several translation studies hypotheses by performing an extensive exploration of the ability of various feature sets to distinguish between O and T. Using SVM classifiers and ten-fold cross-validation evaluation, they listed several features that yield near perfect accuracy.


Most works mentioned above train and evaluate classifiers on texts drawn from the same corpus. When these classifiers are tested on texts from different domains, or in a different genre, or translated from a different language, classification accuracy dramatically deteriorates.
\citet{koppel-ordan:2011:ACL-HLT2011} train classifiers on the Europarl corpus \citep{europarl}, with English translated from five different languages. When the classifiers are evaluated on English translated from the same language they were trained on, accuracy is near~100\%; but when evaluated on translations from a different language, accuracy drops significantly, in some cases below~60\%. This pattern recurs when the test corpus is different from the training corpus (newspaper articles vs.\ parliament proceedings).
Similarly, \citet{udi:llc} report excellent (near~100\%) results identifying Hebrew translationese on a corpus of literary texts, using very simple word-level features. Evaluation on different domains (popular science) and on Hebrew translated from French, rather than English, however, shows much poorer results, with accuracies around~60\% in many cases.

We hypothesize that the main reason for the deterioration in the accuracy of (supervised) translationese classifiers when evaluated out-of-domain stems from the fact that domain differences overshadow the differences between O and T.
\citet{diwersy:etal:2014} studied various sorts of linguistic variation by applying semi-supervised multivariate techniques. They investigated, among other factors,
register variation in English and German originals and translations. By applying a series of supervised and unsupervised statistical analyses, they demonstrated that register-related properties are much better exhibited by the underlying texts than properties related to the documents' translation status.
We address the challenge of mixed-domain classification in Section~\ref{sec:domain-mix}.

One way to overcome the dependence on labeled data and domain-overfitting of supervised classifiers is to use \emph{unsupervised} methods, in particular \emph{clustering}.
The only application of clustering to translationese that we are aware of is the work of \citet{nisioi13}, who investigated translationese- and authorship-related characteristics by applying hierarchical clustering to books written by a Russian-English bilingual author. While they mainly focused on authorship attribution, \citet{nisioi13} also demonstrated that it is possible to discriminate O from T by applying clustering with lexical features (function words) extracted from complete books (25,000--180,000 tokens). We address the challenge of unsupervised identification of translationese using a different methodology and much smaller logical units (2,000 tokens), and further broaden the scope of our work by proposing a methodology for telling O from T in mixed-domain scenarios.

\label{sec:unsupervised}
Unsupervised classification is a well-established discipline;
in this work we use
\emph{KMeans} \citep{Lloyd:2006} for clustering and
\emph{KMeans++} \citep{David-Sergei:2007} as a KMeans initialization method.
%
%
%
%


\section{Experimental Setup}
\label{sec:exp-setup}
\subsection{Datasets}
\label{sec:datasets}
Our main dataset\footnote{The dataset is available at \url{http://cl.haifa.ac.il/projects/translationese}.} consists of texts originally written in English and texts translated to English from French.
We use various corpora:%
\begin{inparaenum}[(i)]
\item Europarl, the proceedings of the European Parliament \citep{europarl}, between the years 2001-2006;
\item the Canadian Hansard, transcripts of the Canadian Parliament, spanning years 2001-2009;
\item literary classics written (or translated) mainly in the 19th century; and
\item transcripts of TED and TEDx talks.
\end{inparaenum}
This collection suggests diversity in genre, register, modality (written vs.\ spoken) and era.
Table~\ref{tbl:corpus-stats} details some statistical data on the corpora (after tokenization).%
\footnote{We use ``EUR'', ``HAN'', ``LIT'' and ``TED'' to denote the four corpora in the discussion below.}
We now briefly describe each dataset.

\begin{table*}[hbt]
\centering
\begin{tabular}{l|rrrrrrr}
& \multicolumn{3}{c}{\textbf{Number of sentences}}
& \multicolumn{2}{c}{\textbf{Number of tokens}}
& \multicolumn{2}{c}{\textbf{Number of types}}
\\ 
\multicolumn{1}{c|}{\textbf{Corpus}} &
\multicolumn{1}{c}{\textbf{Original E}} &
\multicolumn{1}{c}{\textbf{F$\rightarrow$E}} &
\multicolumn{1}{c}{\textbf{Total}} &
\multicolumn{1}{c}{\textbf{Original E}} &
\multicolumn{1}{c}{\textbf{F$\rightarrow$E}} &
\multicolumn{1}{c}{\textbf{Original E}} &
\multicolumn{1}{c}{\textbf{F$\rightarrow$E}}
\\ \hline 
EUR &
134,725 & 71,816 &
206,541 &
3,406,513 & 2,112,085 &
37,203 & 28,119
\\ 
HAN &
3,441,984 & 757,573 &
4,199,557 &
65,491,960 & 13,457,613 &
158,645 & 63,192
\\ 
LIT &
36,123 & 85,210 &
121,333 &
858,297 & 1,750,525 &
25,113 & 38,842
\\ 
TED &
7,551 & 4,827 &
12,378 &
129,334 & 87,214 &
9,667 & 7,441
\\ 
\end{tabular}
\caption{Corpus statistics}
\label{tbl:corpus-stats}
\end{table*}

Europarl is probably the most popular parallel corpus in natural language processing, and it was indeed used for many of the translationese tasks surveyed in Section~\ref{sec:related-work}. This corpus has been used extensively in SMT \citep{koehn-birch-steinberger:2010:MT-Summit-XII}, and was even adapted specifically for research in translation studies: \citet{ISLAM12.729} compiled a customized version of Europarl, where the direction of translation is indicated.
We use a version of Europarl \citep{ella-shuly:corpus:2015} that aims to further increase the confidence in the direction of translation, through a comprehensive cross-lingual validation of the original language of the speakers.

The Hansard is a parallel corpus consisting of transcriptions of the Canadian parliament in English and French between 2001 and 2009. This is the largest available source of English--French sentence pairs. We use a version that is annotated with the original language of each parallel sentence. Relying on  metadata available in the corpus, we filtered out all segments not referring to speech, i.e., retaining only sentences annotated as \emph{Content ParaText}.

The Literature corpus consists of literary classics written (and translated) in the 18th--20th centuries by English and French authors; the raw material is available from the Gutenberg project.
We use subsets that were manually or automatically paragraph-aligned.
Note that classifying literary texts is considered a more challenging task than classifying more ``technical'' translations, such as parliament proceedings, since translators of literature typically enjoy more literary freedom, thereby rendering the translation product more similar to original writing \citep{LynchV12,udi:llc}.

Our TED talks corpus consists of talks originally given in English and talks translated to English from French.
The quality of translations in this corpus is very high: not only are translators assumed to be competent, but the common practice is that each translation passes through a review before being published. This corpus consists of talks delivered orally, but we assume that they were meticulously prepared, so the language is not spontaneous but rather planned. Compared to the other sub-corpora, the TED dataset has some unique characteristics that stem from the following reasons:
\begin{inparaenum}[(i)]
\item its size is relatively small;
\item it exhibits stylistic disparity between the original and translated texts (the former contains more ``oral'' markers of a spoken language, while the latter is a written translation); and finally
\item TED talks are not transcribed but are rather subtitled, so they undergo some editing and rephrasing.%
\footnote{\url{http://translations.ted.org/}}
\end{inparaenum}

The vast majority of TED talks are publicly available online, which makes this corpus easily extendable for future research.

\subsection{Processing and Tools}
\label{sec:proc-and-tools}
All datasets are first tokenized using the Stanford tools \citep{manning-EtAl:2014:P14-5} and then partitioned into chunks of approximately 2000 tokens (ending on a sentence boundary). We assume that translationese-related features are present in the texts across author or speaker, thus we allow some chunks to contain linguistic information from two or more different texts simultaneously.
For the main (single-corpus) classification experiments we use 2000 text chunks each from Europarl and Hansard, 800 from Literature and 88 chunks from TED; each sub-corpus consists of an equal number of original and translated chunks. For every classification experiment we use the maximal equal number of chunks from each class, thus we always (randomly) down-sample the datasets in order to have a comparable number of training/testing examples for supervised classification, and comparable cluster size for clustering.

We use Weka \citep{weka} as the main tool for classification, clustering, and dimension reduction. In all the classification experiments, we use SVM (SMO) as the classification algorithm with the default linear kernel. For clustering we use Weka's KMeans implementation (SimpleKMeans) with the KMeans++ initialization strategy. We use Eucledian distance as the similarity measure for KMeans, and apply a custom clustering-evaluation-based wrapper (see Section~\ref{sec:clustering-results}) to further enhance Weka's basic clustering implementation.

We use Principal Component Analysis (PCA, \citet{Jolliffe:2002}) for dimension reduction. PCA is a statistical procedure that discovers variables with the largest possible variance, i.e., features that account for most variability in the data (\emph{principal components}). It performs a linear mapping of the data to a lower-dimensional space in a way that maximizes the variance of the data in the low-dimensional representation, by removing highly correlated or superfluous variables. The outcome of PCA is a new set of features, each of which is a linear combination of the discovered components. The number of the newly generated variables varies from one to the number of variables originally used to describe the data, and is typically controlled by a parameter.

Apart from the enhanced efficiency (due to the reduced computational costs), dimensionality reduction often carries a positive effect on the accuracy of the underlying classification task, especially when the data are meager or feature vectors are sparse. The (accuracy-wise) optimization gains of PCA, when followed by the KMeans clustering algorithm, were reported by \citet{Ng01onspectral}. We perform dimension reduction using the Weka implementation of PCA, with the ``variance\_covered'' parameter set to 0.1 across all feature types and datasets, prior to applying a clustering procedure.

\subsection{Features}
\label{sec:features}
We focus on a set of features that reflect lexical and structural properties of the text, and have been shown to be effective for supervised classification of translationese \citep{vered:noam:shuly}. Specifically, we use \emph{function words} (FW), more precisely, the same list that was used in previous works on classification of translationese \citep{koppel-ordan:2011:ACL-HLT2011,vered:noam:shuly}.
Feature values are raw counts (further denoted by \emph{term frequency, tf}), normalized by the number of tokens in the chunk; the chunk size may slightly vary, since the chunks respect sentence boundaries. For the clustering experiments we further scale the normalized \emph{tf} by the \emph{inverse document frequency (idf)}, which offsets the importance of a term by a factor proportional to its frequency in the corpus. The \emph{tf-idf} statistic has been shown to be effective with \emph{lexical} features, and is often used as a weighting factor in information retrieval and text mining. While function words are assumed to be very frequent, their counts within a text vary greatly (e.g., ``the'' vs.\ ``whereas''). We therefore opt for \emph{tf-idf} weighting of FW across all sub-corpora.

In addition to function words, we experiment with several other feature sets, including character trigrams, part-of-speech (POS) trigrams, \emph{contextual function words} and \emph{cohesive markers}. Contextual function words are a variation of POS trigrams where a trigram can be anchored by specific function words: these are consecutive triplets $\langle w_1$,$w_2$,$w_3 \rangle$ where at least two of the elements are function words, and at most one is a POS tag. Cohesive markers are words or phrases that signal the underlying flow of thought: they organize a composition of phrases by specifying the type, purpose or direction of upcoming ideas, and can therefore serve as evidence of the translation process. We use the list of~40 cohesive markers defined in \citet{vered:noam:shuly}.

Character, POS, and contextual FW trigrams are calculated as detailed in \citet{vered:noam:shuly}, but we only consider the 1000 most frequent feature values extracted from each dataset (or a combination of datasets) being classified. This subset yields the same classification quality as the full set, reducing computation complexity.


\section{Supervised Classification}
\label{sec:supervised}
We begin with supervised classification, re-establishing the high accuracy of in-domain (supervised) classification of translationese, but highlighting the deterioration in accuracy when cross-domain classification is considered.
We first reproduce the Europarl classification results with the best performing feature sets, as reported by \citet{vered:noam:shuly}, and present results for three additional sub-corpora: Hansard, Literature and TED. Table~\ref{tbl:classification-main} lists the ten-fold cross-validation classification accuracy with various features.
All features (except perhaps cohesive markers) yield excellent accuracy.

\begin{table}[hbt]
\centering
\begin{tabular}{@{}l|rrrr@{}}
\textbf{feature / corpus} & \textbf{EUR} & \textbf{HAN} & \textbf{LIT} & \textbf{TED} \\
\hline 
FW & 96.3 & 98.1 & 97.3 & 97.7 \\
char-trigrams & 98.8 & 97.1 & 99.5 & 100.0 \\
POS-trigrams & 98.5 & 97.2 & 98.7 & 92.0 \\
contextual FW & 95.2 & 96.8 & 94.1 & 86.3 \\
cohesive markers & 83.6 & 86.9 & 78.6 & 81.8 \\
\end{tabular}
\caption{In-domain (cross-validation) classification accuracy using various feature sets}
\label{tbl:classification-main}
\end{table}


A few previous works suggested that cross-domain classification of translationese results in low accuracy \citep{koppel-ordan:2011:ACL-HLT2011,udi:llc}. Our experiments corroborate this observation; Table~\ref{tbl:cross-domain-eh} depicts the cross-domain classification accuracy on the Europarl, Hansard and Literature corpora, when training on one corpus and testing on another (using function words).%
\footnote{We focus mainly on function words, because they are known to reflect stylistic differences rather than contents or specific corpus features, and are therefore less susceptible to domain overfitting. Other feature sets yielded similar results.}
A balanced setup for this experiment was generated by randomly selecting 800 chunks from each corpus, divided equally to O and T. The results only slightly outperform chance level, even for the Europarl--Hansard seemingly domain-related pair: we obtain 59.7\% to 60.8\% accuracy in the two directions.

\begin{table}[hbt]
\centering
\begin{tabular}{@{}l|rrrc@{}}
\textbf{train / test} & \textbf{EUR} & \textbf{HAN} & \textbf{LIT} & \textbf{\splitcellcen{10-fold\\x-validation}} \\
\hline 
EUR & & 60.8 & 56.2 & 94.7 \\ \hline 
HAN & 59.7 & & 58.7 & 98.1 \\ \hline 
LIT & 64.3 & 61.5 & & 97.3 \\
\end{tabular}
\caption{Pairwise cross-domain classification using function words}
\label{tbl:cross-domain-eh}
\end{table}


Attempting to enrich the classifier's training ``experience'' we conducted additional experiments, where we train on two sub-corpora out of Europarl, Hansard and Literature, and test on the remaining one. The results are depicted in Table~\ref{tbl:cross-domain-ehl}. Here, too, accuracy is very low, implying that training on diverse data does not necessarily provide a solution for cross-domain classification of translationese. The right-hand column of the table reports ten-fold cross-validation results of the two sub-corpora that are subject for training. Excellent in-domain classification results on the one hand and poor cross-domain predictive performance on the other, imply that the model describing the relation in a certain domain is inapplicable to a different (even seemingly similar) domain due to significant differences in the distribution of the underlying data.

\begin{table}[hbt]
\centering
\begin{tabular}{@{}p{1.9cm}|p{0.65cm}p{0.65cm}p{0.65cm}c@{}}
\textbf{train / test} & \textbf{EUR} & \textbf{HAN} & \textbf{LIT} & \textbf{\splitcellcen{10-fold\\x-validation}} \\
\hline 
EUR\,+\,HAN & & & 63.8 & 94.0 \\ \hline 
EUR\,+\,LIT & & 64.1 & & 92.9 \\ \hline 
HAN\,+\,LIT & 59.8 & & & 96.0 \\
\end{tabular}
\caption{Leave-one-out cross-domain classification using function words}
\label{tbl:cross-domain-ehl}
\end{table}


Reflecting the poor generalization capability of translationese features, these results call for developing other methodologies for reliably discriminating O from T, specifically, methodologies that are independent of in-domain labeled data.

\section{Clustering}
\label{sec:clustering-results}
\subsection{Initial results}
To overcome the domain-dependence of supervised classification, we experiment in this section with unsupervised methods. We begin with the KMeans clustering algorithm, using KMeans++ initialization policy and dimension reduction. To evaluate the accuracy of the algorithms, each cluster is labeled by the majority of (O or T) instances it includes (using ground truth annotations), and the overall precision is the percentage of instances correctly assigned to their respective clusters (we discuss \emph{unsupervised} cluster labeling in Section~\ref{sec:cluster-labeling}).

The KMeans clustering algorithm (with any initialization policy) is sensitive to the initial settings of its parameters, in particular the initial choice of centroids. A cluster \emph{centroid} is the geometrical center of all observations within the cluster. The result of the KMeans algorithm may significantly vary according to its first step: the initial assignment of (random) points to cluster centroids. We address this potential pitfall by performing $N$ clustering iterations, randomly varying the initial parameter settings, outputting the outcome that exhibits the highest similarity of points within a cluster. Formally, let $C_i^j$ denote cluster $i$ in iteration $j$, and let $\mi^j$ denote this cluster's centroid, so that $i$$\in$[1,2], and $j$$\in$[1..N]. \emph{Sum-of-Square-Error (SSE)} is an intrinsic clustering evaluation metric that measures the similarity of elements in a cluster. The SSE of $C_i^j$ is defined by
\begin{equation*}
SSE_i^j = \sum_{x \in C_i^j}(x-\mi^j)^2
\end{equation*}
We aim to optimize the clustering result by choosing an outcome that minimizes the accumulative SSE:
\[\argmin_j SSE^j =  \argmin_{j \in [1..N]} \sum_{i \in [1,2]} SSE_i^j\] 

The selected clustering outcome represents the result of a \emph{single} clustering experiment. The described method for selecting a clustering outcome can be viewed as a binary version of the \emph{Bisecting} KMeans algorithm; it is applied in all experiments throughout the paper, with number of iterations ($N$) fixed to 5, following the recommendation by \citet[p.~13]{Steinbach00acomparison}.

We conducted a series of experiments with various feature sets; the main results are depicted in Table~\ref{tbl:clustering-main}. The reported numbers reflect the average accuracy over 30 experiments (the only difference being a random choice of the initial conditions).\footnote{Standard deviation in most experiments was close to~0.}

\begin{table}[hbt]
\centering
\begin{tabular}{@{}l|cccc@{}}
\textbf{feature / corpus} & \textbf{EUR} & \textbf{HAN} & \textbf{LIT} & \textbf{TED} \\
\hline 
FW & 88.6 & 88.9 & \textbf{78.8} & \textbf{87.5} \\
char-trigrams & 72.1 & 63.8 & 70.3 & 78.6 \\
POS-trigrams & \textbf{96.9} & 76.0 & 70.7 & 76.1 \\
contextual FW & 92.9 & \textbf{93.2} & 68.2 & 67.0 \\
cohesive markers & 63.1 & 81.2 & 67.1 & 63.0 \\
\end{tabular}
\caption{Clustering results using various feature sets}
\label{tbl:clustering-main}
\end{table}


First and foremost, the results are very good, ranging from a few percent points lower than supervised classification (Table~\ref{tbl:classification-main}, Europarl and Hansard) to approximately~25 percent points lower in a few cases (e.g., Literature). Function words systematically yield very high accuracy; the quality of clustering with other features varies across the sub-corpora. Cohesive markers perform poorly (with a single exception, Hansard), which mirrors the moderate supervised classification precision achieved with the same feature set.

The exceptionally high result of Europarl with POS-trigrams can be attributed to the excessive frequency  of specific phrases in the translated Europarl texts (in contrast to their original counterparts).\footnote{As an example (and in line with \citet{Halteren08}), in the 2000 Europarl chunks, the phrase \textnl{ladies and gentlemen}  appears 1258 times in T, but  only 12 times in O.}
We explain the lower precision achieved on the Literature corpus by its diverse character: it comprises works attributed to a variety of authors, periods and genres, which is challenging for the unsupervised algorithm (see Section~\ref{sec:domain-mix}). A notably high accuracy is obtained on the small TED corpus, which implies the applicability of our clustering methodology to data-meager scenarios.

We conducted an additional set of experiments with unequal proportions of original and translated texts, considering twice the number of O chunks compared to T and vice versa. The average clustering accuracy using FW is similar to that obtained in the balanced setup (Table~\ref{tbl:clustering-main}): 87.5\% on Europarl, 88.9\% on Hansard, 73.2\% on Literature, and 88.6\% on the TED sub-corpus.

\subsection{Cluster labeling}
\label{sec:cluster-labeling}
As is always the case with unsupervised methods, clustering can divide observations into classes but cannot label those classes.
A \emph{cluster labeling} algorithm examines the contents of each cluster in order to find labels that best summarize its members, and distinguish the clusters from each other.

In the context of translationese identification, the task of cluster labeling is to determine which of the produced clusters represents O, and which T. We address this challenge by exploring similarities between the \emph{language models} of the obtained clusters, and language models of (presumably) \emph{prototypical} O and T samples. A simple unigram language model assigns each word a probability proportional to its frequency in the underlying text; we use smoothed term frequencies scaled by the inverse total term frequencies. We then compare language models to reveal similarities between the prototypical O and T samples and the chunk sets produced by clustering.

The construction method of prototypical LMs is motivated by \begin{inparaenum}[(i)]
\item abstracting from content, by utilizing only function words for this purpose; and
\item attempting to avoid the interference of domain-related properties, by considering only (presumably) \emph{universal} markers: words that share similar frequency patterns in several datasets w.r.t.\ to O vs. T. \end{inparaenum}

Let $O_m$ (O-markers) denote a set of function words that tend to be associated with O. We select this set by picking words whose frequency in O is excessive, compared to T; more precisely, the ratio of their frequency in O and T is above~(1+$\delta$), where $\delta$=0.05. Similarly, $T_m$ (T-markers) is a set of words with O-to-T frequency ratio below~(1-$\delta$). We create a prototypical O example by the concatenation of $O_m$, and a prototypical T example by the concatenation of $T_m$. The language model of these examples is then constructed by the $\epsilon$-smoothed likelihood of each term in the markers vocabulary $V=O_m\bigcup T_m$, where $\epsilon$=0.001.

Formally, for $w \in V$,
\begin{equation*}
\label{eq:term-likelihood}
\begin{split}
p(w\mid O_m) &= \frac{\mathit{tf}(w)+\epsilon}{|O_m|+\epsilon\times|V|} \\
p(w\mid T_m) &= \frac{\mathit{tf}(w)+\epsilon}{|T_m|+\epsilon\times|V|}
\end{split}
\end{equation*}

We denote the resulting language models by $P_O$ and $P_T$, respectively. Given two clusters, $C_1$ and $C_2$, we similarly compute their language models, denoted by $P_{C_1}$ and $P_{C_2}$, respectively, over the vocabulary $V$.
We measure the similarity between a class $X$ (either O or T) and a cluster $C_i$ using the Jensen-Shannon divergence (JSD) \citep{Lin:2006} on the respective probability distributions. Specifically, we define the \emph{distance} between the language models as the square root of the divergence value, which is a metric, often referred to as \emph{Jensen-Shannon distance} \citep{EndresS03}:
\begin{equation*}
D_{JS}(X,C_i) = \sqrt[2]{JSD(P_X || P_{C_i})}
\end{equation*}


The assignment of the label $X$ to the cluster $C_1$ is then supported by both $C_1$'s proximity to the class $X$ and \emph{$C_2$}'s proximity to the other class:
\begin{equation*}
\begin{split}
label(C_1)\!=\!
\begin{cases}
    \mbox{``O''} &\mbox{if}\;D_{JS}(O,C_1)\! \times \!D_{JS}(T,C_2)\!<\\
    & \alpha \times D_{JS}(O,C_2)\! \times \!D_{JS}(T,C_1) \\
    \mbox{``T''} &\mbox{otherwise}
\end{cases}
\end{split}
\end{equation*}
\emph{$C_2$} is assigned the complementary label. The value of $\alpha$ is fixed to 1 in this equation, but we note that it can be varied for further investigation of the relatedness of the underlying language models.

We apply the cluster labeling technique described above to determine the labels of generated clusters. We construct prototypical O- and T-texts by selecting O- and T-markers from a random sample of Europarl and Hansard texts, using~600 chunks from each corpus.%
\footnote{This subset of the Europarl and Hansard corpora was used for one-time generation of prototypical O and T language models, and excluded from further use.}
We then compare the language models induced by these samples to those of the generated clusters (tested on different chunks, of course) to determine the cluster labels; the predicted labels are then verified against the majority-driven labeling, based on ground truth annotations.
We apply this procedure to the outcome of all clustering experiments (per domain, using various features), achieving overall precision of 100\%. In other words, the labeling procedure yields prefect accuracy not only on Europarl and Hansard texts that were not used for generation of O and T prototypical examples, but also on unseen Literature and TED datasets. We conclude that it is possible, in general, to determine the labels of clusters produced by our clustering algorithm with perfect accuracy.

\subsection{Clustering consensus among feature sets}
\label{sec:consensus-results}

Since different feature sets have different predictions on our data, we hypothesize that consensus voting can improve the accuracy of clustering. We treat each individual clustering result (based on a certain feature set) as a judge, voting whether a single text chunk belongs to O or to T. We use the cluster labeling method of Section~\ref{sec:cluster-labeling} to determine labels. The final assignment of a label to a cluster is determined by the majority vote of the various judges.

Table~\ref{tbl:cluster-voting} presents the results of these experiments. We compare consensus results to the accuracy achieved by function words, the best-performing single feature set (on average), see Table~\ref{tbl:clustering-main}. Both three judges and five judges yield a consistent increase in accuracy. Five judges systematically (and, on Europarl and Hansard, significantly) outperform the result of clustering with functions words only. This indicates that various features tend to capture different aspects of translationese, that are eventually leveraged by the ``fusion'' of different clustering results into a single, higher-quality outcome.


\subsection{Sensitivity analysis}
\label{sec:sensitivity}
\label{sec:var-num-chunks}
\begin{figure*}[hbt]
\hspace*{-.4in}
\begin{tabular}{cc}
\epsfig{file=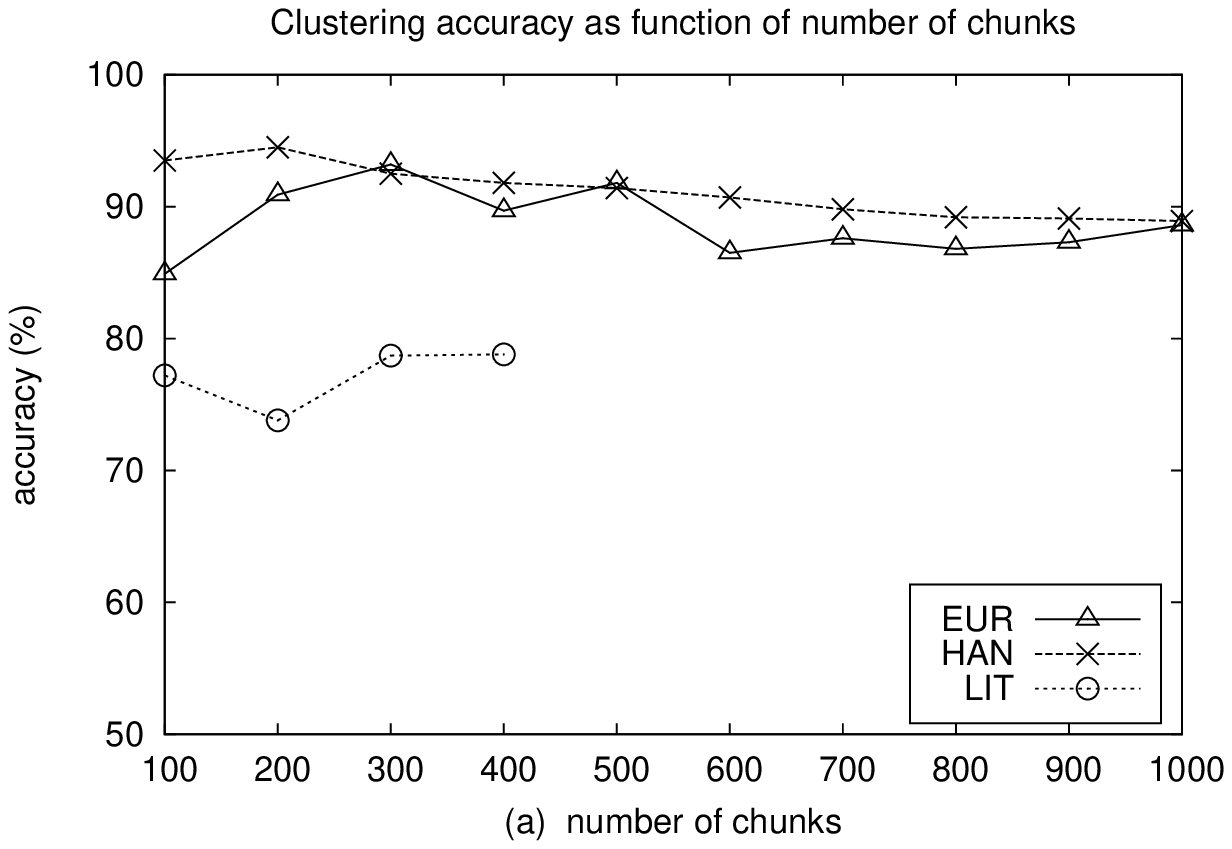,,scale=0.65} & \hspace{-2em}
\epsfig{file=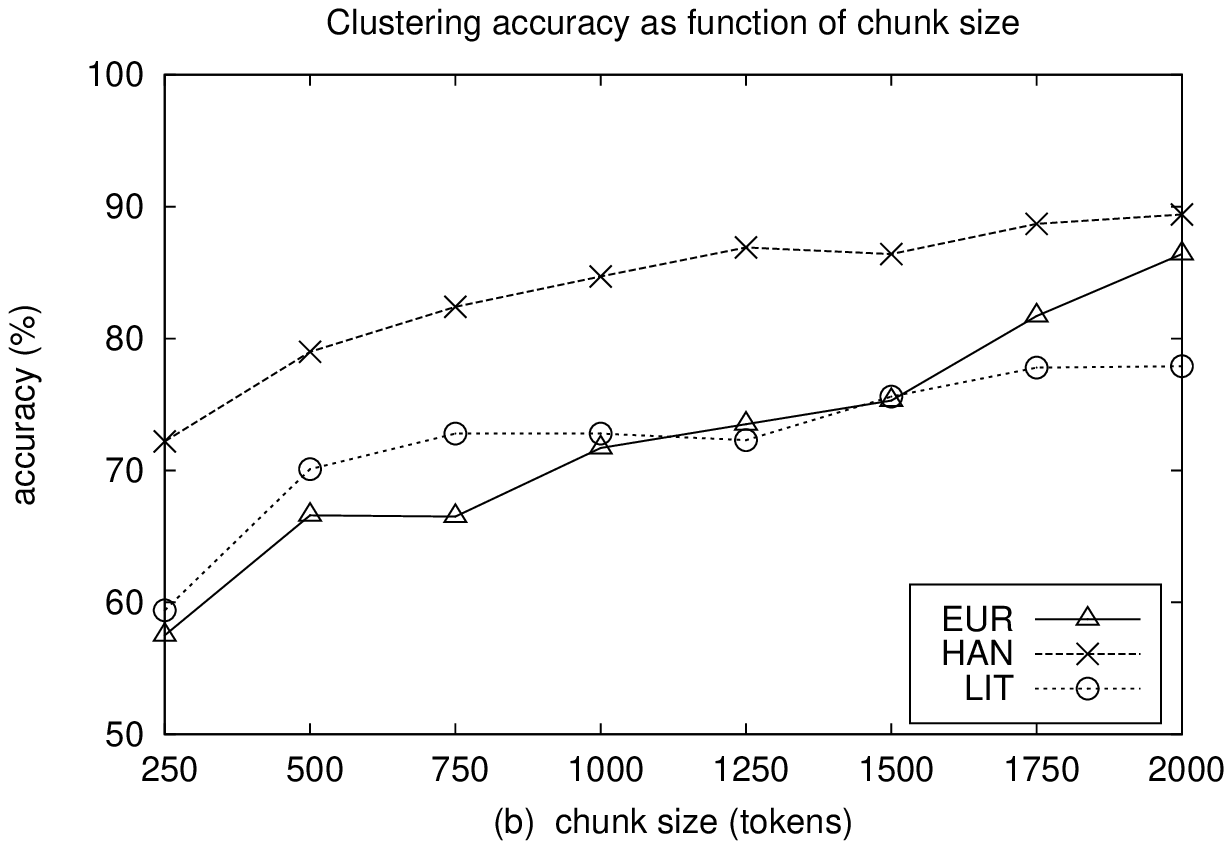,,scale=0.65} 
\end{tabular}
\caption{The effect of varying the number of chunks and chunk size (in tokens) on clustering accuracy}
\label{fig:var-chunk-num-size}
\end{figure*}

In {supervised} classification, the amount of labeled data has a critical effect on the classification accuracy. This does not seem to be the case with clustering: accuracy remains stable when the number of chunks used for classification decreases (Figure~\ref{fig:var-chunk-num-size}a). Evidently, as few as 300 chunks are sufficient for excellent classification.\footnote{The results on the Literature corpus are limited by the amount of available data in this dataset.} We attribute the (slight) fluctuations in the graph to the random choice of the subset of chunks that are subject for clustering. Naturally, clustering accuracy stabilizes when the number of chunks increases, since the effect of random noise diminishes with more data. This result is of clear practical importance, as in real-life situations  only a limited amount of data may be available.


\label{sec:var-chunk-size}
The accuracy of supervised classification deteriorates when the size of the underlying logical units (here, chunks) decreases \citep{Kurokawa:etal:2009}. We corroborate this observation in the context of clustering, but note that reasonable accuracy (over~70\%) can be obtained even with 1000-token chunks (Figure~\ref{fig:var-chunk-num-size}b). This further supports the applicability of unsupervised classification of translationese to real-world scenarios.


%

\begin{table}[hbt]
\centering
\begin{tabular}{@{}l|cccc@{}}
\textbf{method / corpus} & \textbf{EUR} & \textbf{HAN} & \textbf{LIT} & \textbf{TED} \\
\hline 
\splitcellleft{FW} & 88.6 & 88.9 & 78.8 & 87.5 \\ \hline 
\splitcellleft{FW\\char-trigrams\\POS-trigrams} & $91.1^{*}$ & 86.2 & 78.2 & $\textbf{90.9}^{*}$ \\ \hline 
\splitcellleft{FW\\POS-trigrams\\contextual FW} & $\textbf{95.8}^{*}$ & 89.8 & 72.3 & 86.3 \\ \hline 
\splitcellleft{FW\\char-trigrams\\POS-trigrams\\contextual FW\\cohesive markers} & $94.1^{*}$ & $\textbf{91.0}^{*}$ & \textbf{79.2} & 88.6 \\
\end{tabular}
\caption{Clustering consensus by voting; statistically significant improvements, compared to using FW only, are marked with `*'}
\label{tbl:cluster-voting}
\end{table}

\section{Mixed-domain classification}
\label{sec:domain-mix}
Poor cross-domain classification results, as described in Section~\ref{sec:supervised}, demonstrate that the in-domain discriminative features of translated texts cannot be easily generalized to other, even related, domains. In this section we explore the tension between the discriminative power of domain- and translationese-related properties, in the \emph{unsupervised} scenario. Our underlying hypothesis is that domain-specific features overshadow the features of translationese.
The next series of experiments involves (a balanced) combination of various datasets; we excluded the small TED corpus from these experiments to prevent downsampling of other sub-corpora.

\subsection{Domain-related vs.\ translationese-based characteristics}
\label{sec:domain-translationese-char}
We begin with an investigation of the mutual effect of the domain- and translationese-specific characteristics on the accuracy of clustering. We first merged equal numbers of O and T chunks from two corpora: 800 chunks  each from Europarl and Hansard, yielding~1,600 chunks, half of them O and half T. We applied the clustering algorithm of Section~\ref{sec:clustering-results} to this dataset; the result was a perfect domain-driven separation of all Europarl and Hansard chunks, yielding poor (chance-level) translationese accuracy.
In other words, we obtained two clusters, one consisting of Europarl chunks and the other of Hansard chunks, independently of their O-vs.-T status.
We repeated the experiment with additional corpus pairs, and further extended it by adding equal numbers of Literature chunks (400 O and 400 T), this time fixing the number of clusters to three.  Again, the result was separation by domain: Europarl, Hansard and Literature chunks were grouped into distinct clusters (Table~\ref{tbl:clustering-mix}, top).

As an additional experiment, we attempted to leave the decision on the ``best'' number of clusters to the algorithm. To that end, we employed the XMeans clustering procedure \citep{Pelleg-Moore:2000}, which uses KMeans but applies additional statistical cues to decide on the number of clusters that best explain the data. We also applied PCA for dimension reduction prior to XMeans invocation. We repeated both experiments (two- and three-domain mixes) with XMeans, expecting to obtain two and three clusters, respectively. The result is a replication of the more constrained KMeans in three out of four cases 
(Table~\ref{tbl:clustering-mix}, bottom).

These observations have a crucial effect on understanding the tension between the domain- and translationese-based characteristics of the underlying texts. Not only are domains accurately separated given a fixed number of clusters, but even when the decision on the number of clusters is left to the clustering procedure, classification into domains explains the data best (as shown by XMeans). Recall that these experiments all rely on the set of function words: topic-independent features,
that have been proven effective for telling O from T in both supervised (Section~\ref{sec:related-work}) and unsupervised scenarios (Section~\ref{sec:clustering-results}). The fact that this translationese-oriented feature set yields the results presented in Table~\ref{tbl:clustering-mix} clearly demonstrates the dominance of domain-specific properties over the characteristics of translationese.\footnote{Other feature sets yielded similar outcomes.}

\begin{table*}[hbt]
\centering
\begin{tabular}{@{}l|cccc@{}}
\textbf{method / corpus} & \textbf{EUR + HAN} & \textbf{EUR + LIT} & \textbf{HAN + LIT} & \textbf{EUR + HAN + LIT} \\
\hline 
\textbf{KMeans} & & & & \\
accuracy by domain & 93.7 & 99.5 & 99.8 & 92.2 \\
accuracy by translation status & 50.3 & 50.0 & 50.0 & -- \\
\hline 
\textbf{XMeans} & & & & \\
generated \# of clusters & 2 & 2 & 3 & 3 \\
accuracy by domain & 93.6 & 99.5 & 99.9 & 92.2 \\
accuracy by translation status & 50.3 & 50.0 & -- & -- \\
\end{tabular}
\caption{Clustering a chunk-level mix of Europarl, Hansard and Literature using function words; accuracy by translation status (O vs. T) is reported where applicable (i.e., the outcome constitutes two clusters)}
\label{tbl:clustering-mix}
\end{table*}


\subsection{Clustering in a mixed-domain setup}
\label{sec:clustering-domain-mix}
Driven by the results of Section~\ref{sec:domain-translationese-char}, we turn to explore a methodology for identification of translationese in a mixed-domain setup.
We assume that we are given a set of text chunks that come from multiple domains, such that some chunks are O and some are T; the task is to classify the texts to O vs.\ T, \emph{independently of their domain}.
For that purpose, we investigate two approaches: \emph{two-phase} and \emph{flat}. Both methods assume that the number of domains, $k$, is known (it can be discovered by XMeans, as in Section~\ref{sec:domain-translationese-char}, or fixed to a somewhat higher value than estimated in order to capture unsuspected differences within domains). The two-phase method first clusters a mixture of texts into domains (e.g., using KMeans), and then separates each of the resulting (presumably, domain-coherent) clusters into two sub-clusters, presumably O and T. The flat approach applies KMeans, attempting to divide the dataset into $2\times k$ clusters; that is, we expect classification by domains and by translationese status, simultaneously.

\begin{table*}[hbt]
\centering
\begin{tabular}{@{}l|cccc@{}}
\textbf{method / corpus} & \textbf{EUR + HAN} & \textbf{EUR + LIT} & \textbf{HAN + LIT} & \textbf{EUR + HAN + LIT} \\
\hline 
Flat & \textbf{92.5} & 60.7 & 77.5 & 66.8 \\
Two-phase & 91.3 & \textbf{79.4} & \textbf{85.3} & \textbf{67.5} \\
\end{tabular}
\caption{Flat and two-phase clustering of domain-mix using function words}
\label{tbl:flat-hierarchical}
\end{table*}

We experimented with two setups:%
\begin{inparaenum}[(i)]
\item mixture of two datasets out of Europarl, Hansard and Literature (1600 chunks in total); and
\item mixture of all three of them (2400 chunks in total).
\end{inparaenum}
We applied both methods to each of the two setups. We invoked PCA prior to clustering in the flat approach; in the two-phase approach, we applied PCA on \emph{raw} data instances that are subject to clustering at each hierarchy level.\footnote{Note that our two-phase approach differs from the traditional hierarchical clustering in this sense.} As our goal is identification of translationese, we define the accuracy of the classification as the ratio of O and T instances classified correctly (i.e., we ignore the accuracy of identifying the correct domain).


Table~\ref{tbl:flat-hierarchical} reports the results. Both methods yield similarly high accuracy in the Europarl+Hansard setup, and much lower accuracy in the setup of all three datasets (with a single exception of EUR+LIT). This implies that the difficulty of telling O from T increases as the number of domains in the mixed-domain setup grows. The two-phase approach outperforms the flat one in most cases: the latter attempts to cluster data instances by domain and translation status \emph{simultaneously}, and is therefore potentially more error-prone. As a concrete example, in the Europarl+Literature setup, attempting to produce four clusters, we obtained a single cluster of Europarl chunks and three clusters of Literature chunks. The two-phase approach avoids such pitfalls by explicitly separating the steps of domain- and translationese-based clustering.

Table~\ref{tbl:flat-hierarchical} clearly demonstrates that in a real-world scenario, where a dataset can be assumed to include texts from multiple domains, it is possible to overcome the dominance of domain-related features over translationese-related ones by splitting the task into two. The result is highly accurate identification of translated texts, even in an extremely challenging setup. Compare the results of Table~\ref{tbl:flat-hierarchical} to the \emph{supervised} case (Tables~\ref{tbl:cross-domain-eh},~\ref{tbl:cross-domain-ehl}): while clustering cannot compete with ten-fold cross-validation results of heterogenous datasets (93--96\%), it is far superior to training a classifier on one or more datasets and then using it on a data from a new source (60--64\%).


\section{Discussion}
Distinguishing between original and translated texts has been proven useful for SMT, as awareness to translationese can improve the quality of SMT systems. So far, classifying texts into original vs.\ translated has been done almost exclusively by supervised methods. In this work we advocate the use of \emph{unsupervised} classification as an effective way to address this task. We demonstrate that simple feature sets, coupled with standard clustering algorithms, a novel cluster labeling technique, and voting among several features, can yield very high accuracy, over 90\% in several cases. 
Using diverse datasets we robustly demonstrate that the approach we advocate is effective for identification of translationese, even when only little data are available, and text chunks are small. We further highlight the dominance of domain-based characteristics of the texts over their translationese-related properties and propose a simple methodology for identification of translationese in a mixed-domain setup. We conclude that the proposed (two-phase) clustering approach is a robust method for distinguishing O from T in heterogenous datasets.

By conducting a series of experiments with unbalanced proportions of O and T texts, we demonstrate that the proposed methodology is also applicable to scenarios where the original and translated data are unevenly distributed.

We applied PCA for dimension reduction and the \emph{tf-idf} weighting scheme with FW throughout all experiments in this work. The latter had a slight positive effect on clustering accuracy in most scenarios, and no impact in some  cases. Dimension reduction improved computational efficiency, especially with large feature sets (e.g., character and POS trigrams). However, its effect on clustering accuracy was not uniform: the most prominent improvement (over 15 percent points) was obtained on the TED dataset, while a slight accuracy deterioration was observed in a few cases (e.g., 5 percent points on Europarl with FW). We conclude that while carrying an overall positive value, the application of dimension reduction in similar scenarios calls for further investigation.

\section{Conclusion}
To the best of our knowledge, this is the first work to extensively explore unsupervised classification of translationese. We only scratched the surface of this research direction. In the future, we intend to explore the robustness of our approach even further, with more datasets in various language pairs. We will first attempt to identify translationese in \emph{French}, using the current dataset (in the reverse direction). We will also experiment with English-German, in both directions, and hopefully also with English-Hebrew, a more challenging setup.

The potential value of unsupervised identification of translationese leaves much room for further exploratory activities. Our future plans include using various datasets and reduced amount of data for LMs compiled for cluster labeling; in particular, we plan to explore the correlation between these two parameters and the scaling factor $\alpha$ used for association of a label with a clustering outcome.

Furthermore, to highlight the contribution of these results to SMT, we plan to replicate the results of \citet{lembersky-ordan-wintner:CL2012,lembersky-ordan-wintner:CL2014}, using \emph{predicted} rather than ground-truth indication of the translationese status of the texts that are used to train SMT systems. We believe that we will be able to show an improvement in the quality of SMT with extremely little supervision. 

\subsection*{Acknowledgements}
This research was supported by a grant from the Israeli Ministry of Science and Technology. We are grateful to Cyril Goutte, George Foster and Pierre Isabelle for providing us with an annotated version of the Hansard corpus; and to Andr\'{a}s Farkas\footnote{\url{http://farkastranslations.com}} and Fran\c{c}ois Yvon for providing us with the Literary corpus.
Finally, we are indebted to Noam Ordan, Tamir Hazan, Haggai Roitman and Ekaterina Lapshinova-Koltunski for commenting on earlier versions of this article.
All remaining errors and misconceptions are, of course, our own. 

\bibliographystyle{plainnat}
\bibliography{all}

\end{document}